\documentclass[10pt,twocolumn,letterpaper]{article}

\usepackage[pagenumbers]{wacv} 

\usepackage{graphicx}
\usepackage{amsmath}
\usepackage{amssymb}
\usepackage{booktabs}
\usepackage{caption}
\usepackage{multirow}
\usepackage{pifont}
\usepackage{float}
\usepackage[T1]{fontenc}
\usepackage{multirow}
\usepackage[table,xcdraw]{xcolor}
\usepackage{subcaption}

\usepackage{mathtools}

\DeclarePairedDelimiterX{\infdivx}[2]{(}{)}{%
  #1\;\delimsize\|\;#2%
}
\newcommand{\dkl}{D_{KL}\infdivx}

\usepackage[pagebackref,breaklinks,colorlinks]{hyperref}

\usepackage[capitalize]{cleveref}
\crefname{section}{Sec.}{Secs.}
\Crefname{section}{Section}{Sections}
\Crefname{table}{Table}{Tables}
\crefname{table}{Tab.}{Tabs.}

\def\wacvPaperID{*****} 
\def\confName{WACV}
\def\confYear{2024}

\begin{document}

\title{Diffused Heads: Diffusion Models Beat GANs on Talking-Face Generation}

\author{Micha{\l} Stypu{\l}kowski$^1$\\
{\tt\small michal.stypulkowski@cs.uni.wroc.pl}
\and
Konstantinos Vougioukas$^2$ \\
{\tt\small k.vougioukas@imperial.ac.uk}
\and
Sen He \\
{\tt\small senhe752@gmail.com}
\and
Maciej Zi\k{e}ba$^{3, 4}$ \\
{\tt\small maciej.zieba@pwr.edu.pl}
\and
Stavros Petridis$^2$ \\
{\tt\small sp104@imperial.ac.uk}
\and
Maja Pantic$^2$ \\
{\tt\small m.pantic@imperial.ac.uk}
\and 
$^1$University of Wroc{\l}aw
\and 
$^2$Imperial College London
\and
$^3$Wroc{\l}aw University of Science and Technology
\and
$^4$Tooploox
}


\twocolumn[{%
\renewcommand\twocolumn[1][]{#1}%
\maketitle
\begin{center}
    \centering
    \captionsetup{type=figure}
    \includegraphics[width=.8\textwidth]{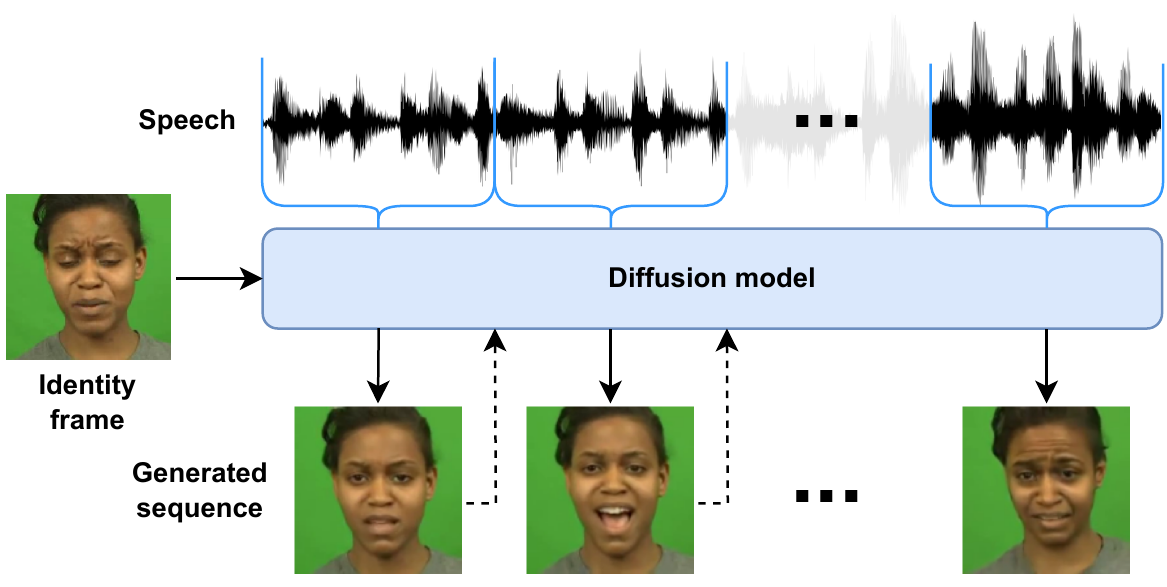}
    \captionof{figure}{Overview of the proposed approach. Given a single identity frame and an audio clip containing speech, the model uses a diffusion model to sample consecutive frames in an autoregressive manner, preserving the identity, and modeling lip and head movement to match the audio input. Contrary to other methods, no additional guidance is required.}
\end{center}%
}]

\begin{abstract}
   Talking face generation has historically struggled to produce head movements and natural facial expressions without guidance from additional reference videos. Recent developments in diffusion-based generative models allow for more realistic and stable data synthesis and their performance on image and video generation has surpassed that of other generative models. In this work, we present an autoregressive diffusion model that requires only one identity image and audio sequence to generate a video of a realistic talking head. Our solution is capable of hallucinating head movements, facial expressions, such as blinks, and preserving a given background. We evaluate our model on two different datasets, achieving state-of-the-art results in expressiveness and smoothness on both of them.\footnote{Project page: \href{https://mstypulkowski.github.io/diffusedheads/}{https://mstypulkowski.github.io/diffusedheads/}.}
\end{abstract}

\section{Introduction}
 
Animation of faces from speech can have a broad scope of applications from an alternative to video compression during virtual calls with poor connectivity, to artistic animation for entertainment industry applications, \eg movies, video games, and VR experience.
Up to date, existing methods struggle to create naturally-looking faces that maintain genuine expressions and movements, while still requiring additional supervision during the generation process.

Deep generative models are constantly gaining popularity and achieving impressive results in image and video generation tasks and have become the \textit{defacto} standard for most facial animation systems.
In particular speech-driven facial animation systems, which are a simple and effective way of producing character animations, have been revolutionized by the introduction of recent generative models such as Generative Adversarial Networks (GANs) \cite{goodfellow2020generative}. GANs are known for being able to produce high-quality frames while simultaneously giving a large degree of control over the generation process \cite{Patashnik_2021_ICCV, karras2020analyzing, pumarola2018ganimation}.

Despite the powerful capabilities of GANs, their application to speech-driven video synthesis has several drawbacks. Firstly, GAN training is notoriously difficult, often requiring an extensive architectural search and parameter tuning to achieve convergence.  The training stability of GAN-based facial animation methods can be improved through the use of additional guidance such as masks or driving frames to guide the generation process. However, this limits them to applications of facial reenactment and reduces their ability to produce original head motions and facial expressions. Furthermore, GAN training can often lead to mode collapse, \ie a situation when the generator can't produce samples that cover the entire support of the data distribution and instead learns to generate only a few unique samples \cite{arjovsky2017towards}.
Finally, existing one-shot GAN-based solutions have problems with face distortion in the generated videos, especially when generating videos with large head motions. This is often solved by either switching to a few-shot approach (\ie using several frames or a short clip) or relying on pre-trained face verification models that serve as oracles for maintaining identity consistency. 

We address all of the above problems, proposing Diffused Heads - a frame-based diffusion model that produces realistic videos, requiring only one identity frame and a speech recording. Generated heads move and behave in a natural expressive way while still preserving the subject's identity and plausible lip sync. In contrast to most recent approaches \cite{vougioukas2020realistic, prajwal2020lip, zhou2020makelttalk, chen2020talking, song2022everybody, zhou2021pose, ren2021pirenderer, ji2022eamm, yin2022styleheat, lahiri2021lipsync3d}, we use Denoising Diffusion Probabilistic Models \cite{ho2020denoising, nichol2021improved} that utilize a variational approach instead of adversarial training and do not require stabilizing discriminators \cite{vougioukas2020realistic, prajwal2020lip, zhou2020makelttalk, zhou2021pose}. To eliminate the problem of unnaturally-looking sequences, we introduce motion frames (see Section \ref{sec:motion_frames}) that are guiding video creation. To maintain the consistency between the speech and the generated frames, we postulate to use audio embeddings extracted from a pre-trained temporal model injected into the model via our novel 
conditioning approach. Finally, instead of using a pre-trained oracle model, we introduce a simple modification of the loss function to preserve the consistency of the lip movement.

Contributions of our work are summarized as follows: 1{)} To the best of our knowledge, we present the first solution for talking-face generation based on diffusion models. 2{)} We enrich the diffusion model with motion frames and audio embeddings in order to maintain the consistency of generated images. 3{)} Our approach is robust in terms of generalization, invariant on the source of identity frames and audio recordings.

\section{Related work}

The problem of speech-driven video synthesis was initially investigated in \cite{yehia1998quantitative}, where the authors discovered a strong correlation between acoustic and video features. Some of the earliest approaches utilized Hidden Markov Models (HMMs) to capture the dynamics of the video and speech sequences \cite{simons1990generation, yamamoto1998lip, xie2007coupled}. The authors of \cite{simons1990generation} used the compact feature representations of speech and video jointly as states of the fully-connected Markov model. In \cite{yamamoto1998lip} the authors used HMMs to estimate the sequence of lip parameters. The authors of \cite{xie2007coupled} proposed a coupled hidden Markov model (CHMM) approach to video-realistic speech animation, which realizes realistic facial animations driven by speaker-independent continuous speech.

Following the modern trends in machine learning, deep learning approaches gained the most promising results in the audio-based video synthesis domain. In \cite{taylor2017deep} the authors propose to use a deep learning model that learns arbitrary nonlinear mappings from phoneme label input sequences to mouth movements in a way that accurately captures natural motion and visual coarticulation effects. In \cite{karras2017audio} a convolutional network is used to transform audio features to 3D meshes of a specific person. The authors of \cite{Siarohin_2019_NeurIPS} presented unsupervised keypoint detection and warping for a motion transfer. Several approaches explore the variations of recurrent models \cite{fan2015photo, suwajanakorn2017synthesizing, pham2017speech, zhou2020makelttalk}.

The most up-to-date approaches for speech-driven video synthesis are based on generative models. Variations of Generative Adversarial Networks (GANs) \cite{goodfellow2020generative} were primarily applied to the problem of video generation \cite{pumarola2018ganimation, pham2018generative, tulyakov2018mocogan}. The GAN-based approach for a speech-driven generation was introduced in \cite{vougioukas2020realistic}. The authors propose an end-to-end system that generates videos of a talking head, using only a still image of a person and an audio clip containing speech without relying on handcrafted intermediate features. They achieve it by utilizing temporal GAN that uses three discriminators focused on achieving detailed frames, audio-visual synchronization, and realistic expressions. In \cite{prajwal2020lip}, the authors propose to incorporate an additional pre-trained Lip-Sync expert during the training to maintain the consistency of generated videos. The paper \cite{chen2020talking} introduces a 3D-aware generative network along with a hybrid embedding module that assures rhythmic head motion. The model presented in \cite{zhou2021pose} modularizes audio-visual representations by devising an implicit low-dimension pose code to tackle the problem of rhythmic head motion. In StyleHEAT \cite{yin2022styleheat}, the authors show how to utilize StyleGAN \cite{karras2020analyzing} model to create talking faces guided by speech embeddings but also controlled by intuitive or attribute editing.  

Some modern approaches utilize rendering networks to obtain more accurate face 3D representation. In \cite{song2022everybody} the authors introduce a novel video rendering network and a dynamic programming method to construct a temporally coherent and photo-realistic video. The authors of \cite{guo2021ad} propose to use an audio-conditioned implicit function to generate a dynamic neural radiance field, from which a high-fidelity talking-head video corresponding to the audio signal is synthesized using volume rendering. A Portrait Image Neural Renderer (PIRenderer) is introduced in  \cite{ren2021pirenderer} that controls the face motions with the parameters of a three-dimensional morphable face model. In \cite{ji2022eamm}, Implicit Emotion Displacement Learner, together with Dense Warping Field, are used to obtain high-quality images. 

The Denoising Diffusion Probabilistic Models \cite{ho2020denoising} are gaining popularity and often outperforming GANs in tasks like image synthesis \cite{dhariwal2021diffusion}, and other guided image generation tasks \cite{rombach2022high, ramesh2022hierarchical, nichol2021glide, radford2021learning}. 
Several attempts utilize that group of models in video generation \cite{harvey2022flexible, hoppe2022diffusion, ho2022video, singer2022make, ho2022imagen}.

To the best of our knowledge, there are no direct attempts to solve speech-driven video synthesis problems using diffusion models. 
Moreover, our method is the first one-shot approach that can hallucinate diverse head motions and does not require an actor to drive the movement via additional visual guidance input. The realism of the gestures is on par with or superior to that of facial reenactment methods.

\section{Diffusion models}

\begin{figure}[t]
  \centering
    \includegraphics[width=0.99\linewidth]{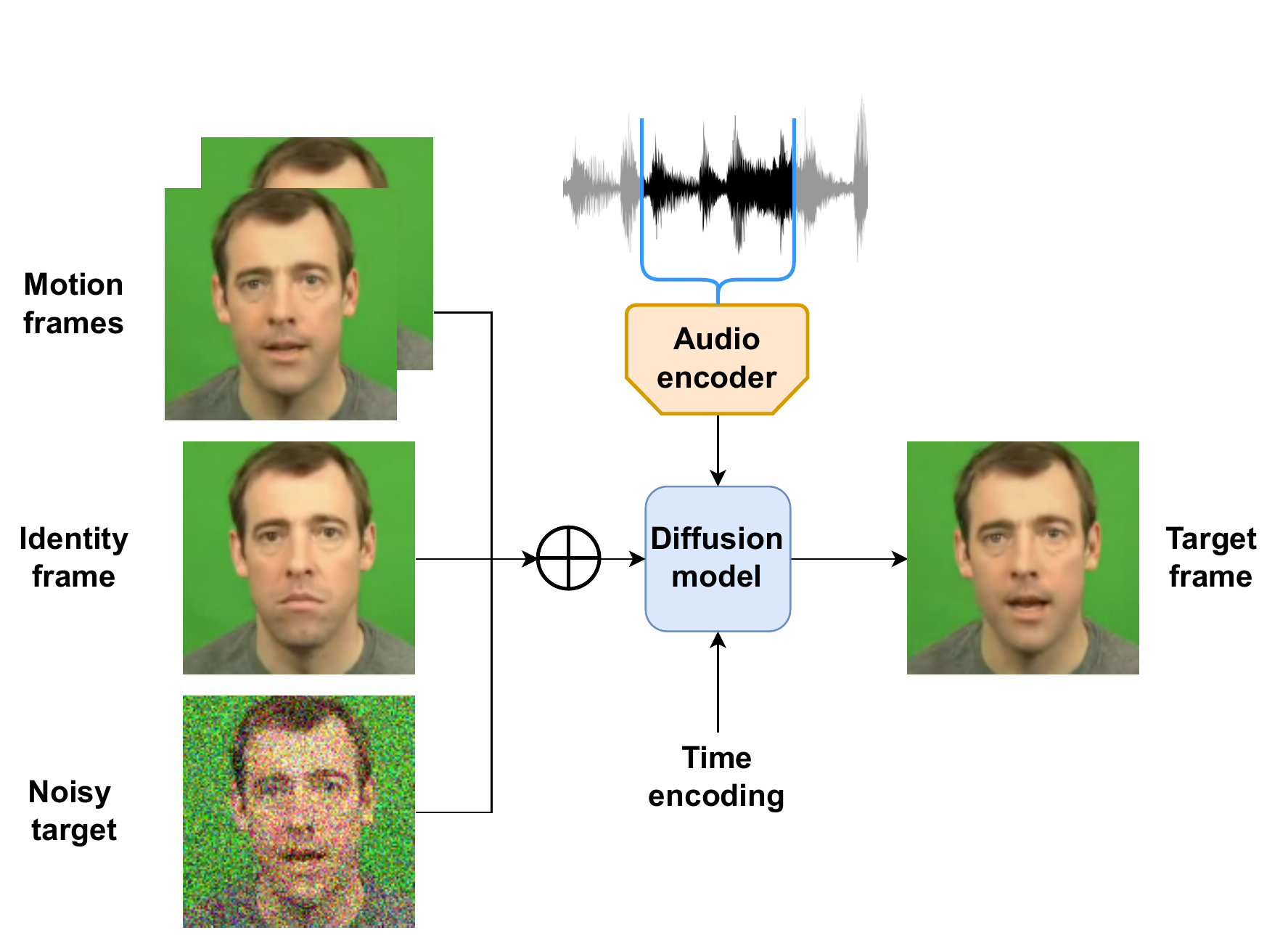}
   \caption{Training step of Diffused Heads. Our model learns to denoise one frame at a time, using identity and motion frames, and an audio embedding extracted from a pre-trained audio encoder. The identity frame informs the model what the face of interest is, and the motion frames are utilized to preserve the movement.}
   \label{fig:model}
\end{figure}


Let us assume we are given samples $x_0$ from a data distribution $x_0 \sim q(x_0)$. We can define a \textit{forward process} $q(x_{1:T}|x_0) := \prod_{t=1}^T q(x_t|x_{t-1})$ that gradually adds Gaussian noise to the data:
\begin{equation}
    q(x_t|x_{t-1}) := \mathcal{N}(x_t; \sqrt{1 - \beta_t} x_{t-1}, \beta_t \mathbf{I})
\end{equation}
where $T$ defines a number of diffusion steps, and $\{ \beta_t \}_{t=1}^T$ is a noise schedule starting from low values that increase with $t$. Note that $\{ \beta_t \}_{t=1}^T$ is known beforehand, so the entire forward process is fixed. For sufficiently large $T$ and $\beta_t \in (0, 1)$, $x_T$ becomes close to a sample drawn from isotropic Gaussian distribution, \ie $x_T \sim \mathcal{N}(0, \mathbf{I})$.

An interesting property of the forward process is the fact that one can access its intermediate states in only one step, that is: 
\begin{equation}
\label{eq:forward_property}
    q(x_t | x_0) = \mathcal{N}(x_t; \sqrt{\bar{\alpha}_t} x_0, (1-\bar{\alpha}_t) \mathbf{I})    
\end{equation} 
where $\bar{\alpha}_t = \prod_{s=1}^t \alpha_s$, and $\alpha_t = 1 - \beta_t$. It allows us to train the model more efficiently.

We are interested in learning how to denoise a sample drawn from Gaussian distribution back to the data. Note that $q(x_{t-1} | x_t)$ depends on the entire dataset and hence is intractable. Intuitively, learning how to denoise $x_T$ into the underlying data point is only possible when the model is given explicit information about where the forward diffusion process started for it. Thus, we can additionally condition $q(x_{t-1} | x_t)$ on $x_0$, making it tractable. Using Bayes' theorem we get:
\begin{equation}
    q(x_{t-1} | x_t, x_0) = \mathcal{N}(x_{t-1}; \tilde{\mu}(x_t, x_0), \tilde{\beta}_t \mathbf{I})
\end{equation}
where:
\begin{align}
    \tilde{\beta}_t &:= \frac{1 - \bar{\alpha}_{t-1}}{1 - \bar{\alpha}_{t}} \beta_t \\
    \tilde{\mu}(x_t, x_0) &:= \frac{\sqrt{\bar{\alpha}_{t-1}}\beta_t}{1 - \bar{\alpha}_t} x_0 + \frac{\sqrt{\alpha_t}(1 - \bar{\alpha}_{t-1})}{1 - \bar{\alpha}_{t}} x_t
\end{align}

Similarly to VAE framework, we define variational posteriors that approximate $q(x_{t-1} | x_t, x_0)$:
\begin{equation}
\label{eq:variational_posterior}
    p_{\theta}(x_{t-1} | x_t) := \mathcal{N}(x_{t-1}; \mu_{\theta}(x_t, t), \Sigma_{\theta}(x_t, t))
\end{equation}

As in \cite{nichol2021improved}, $\mu_{\theta}(x_t, t)$ and $\Sigma_{\theta}(x_t, t))$ are further reparameterized into:
\begin{align}
    \mu_{\theta}(x_t, t) &= \frac{1}{\sqrt{\alpha_t}} \bigg(x_t - \frac{\beta_t}{\sqrt{1 - \bar{\alpha}_t}}\epsilon_{\theta}(x_t, t) \bigg) \\
    \Sigma_{\theta}(x_t, t)) &= \exp(\nu \log \beta_t + (1 - \nu) \log  \tilde{\beta}_t)
\end{align}
where $\epsilon_{\theta}(x_t, t)$ is the model's prediction of the Gaussian noise $\epsilon$ applied on $x_0$ in the process of getting $x_t$, and $\nu$ is an additional output of the model. Nichol \& Dhariwal in \cite{nichol2021improved} proposed to train $\mu_{\theta}$ and $\Sigma_{\theta}$ separately, using $L_{simple}$ and $L_{vlb}$ respectively, where:
\begin{align}
\label{eq:l_simple}
    L_{simple} &:= \mathbb{E}_{t, x_0, \epsilon} \big[||\epsilon - \epsilon_{\theta}(x_t, t) || ^ 2\big]
\end{align}
and $L_{vlb}$ is the variational lower bound (VLB) defined as:
\begin{align}
\label{eq:l_vlb}
    L_{vlb} := L_0 + L_1 + \dots + L_{T-1} + L_T
\end{align}
where:
\begin{align}
    L_0 &:= -\log p_{\theta}(x_0 | x_1) \\
    L_t &:= \dkl{q(x_t|x_{t+1}, x_0)}{p_{\theta}(x_t | x_{t+1})} \\
    & \hspace{8em} \text{ for } t \in \{1, \dots, T-1\} \notag \\
    L_T &:= \dkl{q(x_T|x_0)}{p_{\theta}(x_T)}
\end{align}
For images, $L_0$ is a discretized Gaussian distribution as proposed in \cite{ho2020denoising}. $L_T$ is omitted because $q$ has no trainable parameters and $p_{\theta}(x_T)$ is a Gaussian prior. All of the other terms are Kullback–Leibler divergences between two Gaussian distributions that can be written in a closed form.

In practice, a 2D UNet \cite{ronneberger2015u} with skip-connections, and attention layers is used as a backbone to predict both noise $\epsilon_{\theta}(x_t, t)$ and variance $\Sigma_{\theta}(x_t, t)$. Information about timestep $t$ is injected using corresponding time embedding $\psi(t)$ and group normalization (GN):
\begin{equation}
\label{eq:gn_conditioning}
    h_{s + 1} = t_s \text{GN}(h_s) + t_b
\end{equation}
where $h_s$ and $h_{s+1}$ are consecutive hidden states of UNet, and $(t_s, t_b) = \text{MLP}(\psi(t))$, where MLP is a shallow neural network consisting of linear layers.

\section{Method}
\label{sec:method}

\begin{figure}[t]
  \centering
    \includegraphics[width=0.80\linewidth]{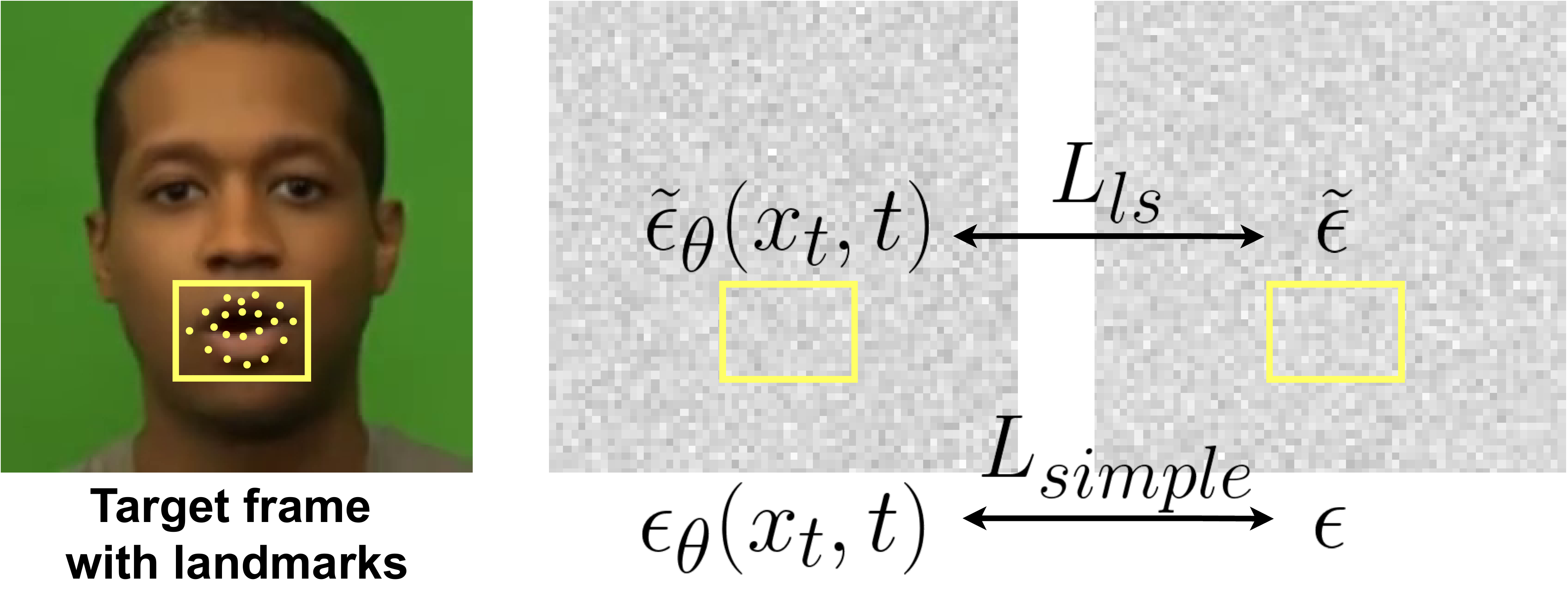}
   \caption{In addition to minimizing L2 distance between ground truth noise $\epsilon$ and predicted noise $\epsilon_{\theta}(x_t, t)$ in $L_{simple}$, we utilize the target frame's landmarks to minimize lip sync loss $L_{ls}$ between cropped ground truth noise $\Tilde{\epsilon}$ and corresponding predicted area $\Tilde{\epsilon}_{\theta}(x_t, t)$.}
   \label{fig:lip_loss}
\end{figure}


\begin{figure*}[t]
  \centering
    \includegraphics[width=0.9\linewidth]{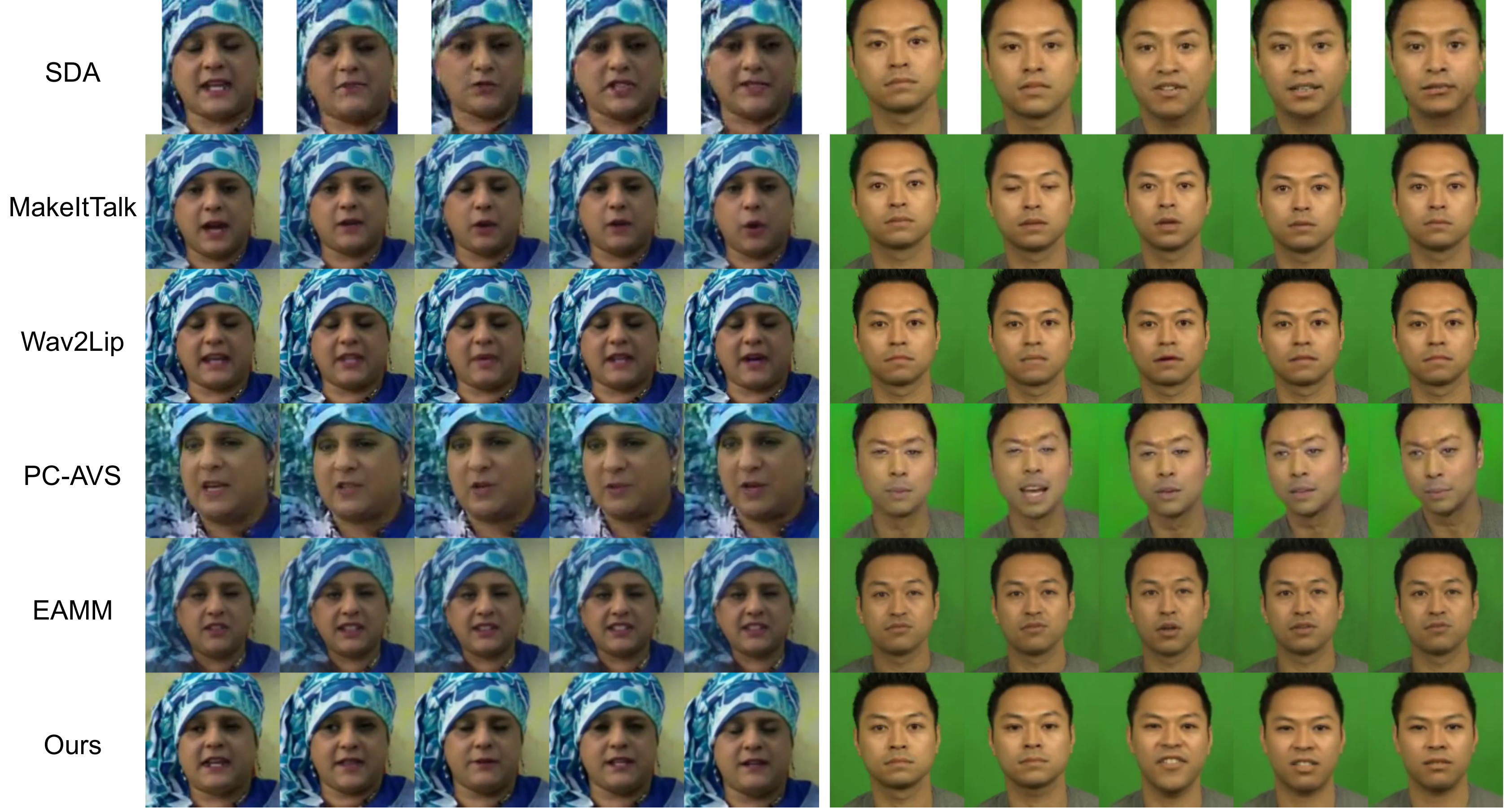}
  \caption{Comparison with other methods on LRW \cite{chung2016lip} (left) and CREMA \cite{cao2014crema} (right) datasets.}
  \label{fig:qual_results}
\end{figure*}

Diffused Heads generates one frame at a time given an identity frame that stays fixed during the entire generation process, and a speech recording embedded using a pre-trained audio encoder. To achieve smoother and more expressive results, we inject additional information on past movement and future expressions by motion frames (Section \ref{sec:motion_frames}) and audio embeddings (Section \ref{sec:speech}). Moreover, an additional lip sync loss (Section \ref{sec:lip_loss}) is defined to force the model to pay more attention to the mouth region.

\subsection{Training}
We train a diffusion model to learn the distribution of frames extracted from videos. The training process is shown in Figure \ref{fig:model}. We randomly sample a video $X = \{x^{(1)}, \dots, x^{(n)}\}$ from the training set, and then a frame $x^{(k)}$ from $X$. $n$ is the total number of frames. In addition to the standard diffusion model's inputs, \ie a time step $t$ and the frame with added noise $x^{(k)}_t$ (following Equation \eqref{eq:forward_property}), to keep the actor's identity, we concatenate $x^{(k)}_t$ with an \textit{identity frame} $x_{Id}$ channel-wise:
\begin{equation}
\label{eq:concat_init}
    x_{In, t}^{(k)} := x^{(k)}_t \oplus_c x_{Id}
\end{equation}
$x_{Id}$ is randomly chosen from $X$. Selecting the identity frame randomly instead of $x^{(0)}$ during the training makes the model familiar to a larger variety of frames as input. In consequence, the generation's robustness is improved.

To add temporal information, we split the corresponding audio sequence into chunks of equal length based on the number of frames in the video. Then, using an audio encoder from \cite{vougioukas2020realistic} pre-trained on the LRW \cite{chung2016lip} dataset, the audio chunks are encoded into audio embeddings $Y = \{y^{(1)}, \dots, y^{(n)}\}$. The details of our proposed audio conditioning method can be found in Section \ref{sec:speech}.

\subsection{Motion frames}
\label{sec:motion_frames}
Even though temporal information is provided to the model by the audio encoder, it is not enough to generate smooth videos. To overcome this problem and preserve the motion, for the target frame $x^{(k)}$ we introduce \textit{motion frames} $x_{Motion}^{(k)} = \bigoplus_c(\{x^{(k-m_x)}, \dots, x^{(k-1)}\})$, where $m_x$ is the number of motion frames, and $\bigoplus_c(.)$ is concatenation operation in channel dimension on all of its arguments. During our ablation study (Section \ref{sec:ablation}), we found that the best value for $m_x$ is 2.

If there are not enough frames preceding $x^{(k)}$, the most natural choice is to fill the remaining motion frames with duplicates of $x^{(0)}$. However, during sampling, we have no access to any ground truth frames, except the identity one. We also do not necessarily want the generated video to start with an exact facial expression as given in the identity frame, \eg when the audio recording starts with silence and the person in the identity frame has their mouth open. Thus, to make the model robust on sample initialization, we utilize $x_{Id}$ as a substitute for missing motion frames.

The motion frames are added to Equation \eqref{eq:concat_init} and get the final form of the direct input to the model:
\begin{equation}
    x_{In, t}^{(k)} := x^{(k)}_t \oplus_c x_{Id} \oplus_c x_{Motion}^{(k)}
\end{equation}

\subsection{Speech conditioning}
\label{sec:speech}
We propose to inject information from audio embedding $y^{(k)}$ by modifying Equation \eqref{eq:gn_conditioning} into:
\begin{equation}
\label{eq:gn_full}
    h_{s + 1} = y^{(k)}_s(t_s \text{GN}(h_s) + t_b) + y_b^{(k)}
\end{equation}
where $(y^{(k)}_s, y^{(k)}_b) = \text{MLP}(y^{(k)})$. In this setting, we shift and scale hidden states of the UNet with the information not only from the time encoding but the audio embedding as well. We found this approach works better in comparison to other conditioning methods, such as using just an additional scale on top of Equation \eqref{eq:gn_conditioning} \cite{preechakul2022diffusion}, and applying a multi-head attention mechanism with queries being a function of the audio embedding \cite{rombach2022high}.

In contrast to motion frames which during sampling are only available for already processed frames, we have an access to the entire speech recording beforehand. To make use of it, we introduce \textit{motion audio embeddings} that bring information from both past and future audio segments. We define them as a vector created by concatenating selected audio embeddings: $y_{Motion}^{(k)} = \bigoplus(\{y^{(k-m_y)}, \dots, y^{(k)}, \dots, y^{(k+m_y)}\})$, where $m_y$ is the number of additional audio embeddings from one side. The details of our choice of $m_y$ can be found in the ablation study in Section \ref{sec:ablation}. Similarly to motion frames, if we run out of embeddings, we pad $y_{Motion}^{(k)}$ with either $y^{(0)}$ at the beginning or $y^{(n)}$ at the end. Finally, we use $y_{Motion}^{(k)}$ instead of $y^{(k)}$ in Equation \eqref{eq:gn_full}.

\subsection{Lip sync loss}
\label{sec:lip_loss}


\begin{table*}[]
\centering
\begin{tabular}{ccccccccccc}
                        &            & FVD $\downarrow$                                    & FID $\downarrow$                                   & Blinks/s                            & Blink dur.                            & OFM                                    & F-MSE                                  & AV off.                            & AV Conf. $\uparrow$      & WER $\downarrow$                                  \\ \hline
                        & SDA        & 198.84                                 & 61.95                                  & \cellcolor[HTML]{B6D7A8}\textbf{0.52} & \cellcolor[HTML]{B6D7A8}\textbf{0.28} & \cellcolor[HTML]{D9EAD3}73.82          & \cellcolor[HTML]{D9EAD3}18.94          & \cellcolor[HTML]{B6D7A8}\textbf{1} & \cellcolor[HTML]{B6D7A8}\textbf{7.40} & 0.77                                  \\
                        & MakeItTalk & 269.29                                 & 7.57                                   & 0.09                                  & \cellcolor[HTML]{B6D7A8}\textbf{0.28} & 57.21                                  & 3.44                                   & -3                                 & 3.16                                  & 0.99                                  \\
                        & Wav2Lip*    & 366.14                                 & \cellcolor[HTML]{B6D7A8}\textbf{2.83}  & 0.03                                  & 0.16                                  & 47.12                                  & 1.45                                   & \cellcolor[HTML]{D9EAD3}-2         & \cellcolor[HTML]{D9EAD3}6.58          & \cellcolor[HTML]{B6D7A8}\textbf{0.51} \\
                        & PC-AVS     & \cellcolor[HTML]{D9EAD3}153.12         & 11.96                                  & 0.20                                  & 0.16                                  & 69.59                                  & 17.13                                  & -3                                 & 6.24                                  & \cellcolor[HTML]{D9EAD3}0.64          \\
                        & EAMM       & 172.18                                 & 9.28                                   & 0.03                                  & 0.16                                  & 58.46                                  & 4.39                                   & -3                                 & 3.83                                  & 0.95                                  \\
                        & Ours       & \cellcolor[HTML]{B6D7A8}\textbf{71.88} & \cellcolor[HTML]{D9EAD3}3.94           & \cellcolor[HTML]{D9EAD3}0.35          & \cellcolor[HTML]{B6D7A8}\textbf{0.28} & \cellcolor[HTML]{B6D7A8}\textbf{70.71} & \cellcolor[HTML]{B6D7A8}\textbf{19.69} & \cellcolor[HTML]{D9EAD3}-2         & 4.61                                  & 0.77                                  \\
\multirow{-7}{*}{\rotatebox[origin=c]{90}{LRW}} & GT         &                                        &                                        & {\color[HTML]{B7B7B7} 0.53}           & {\color[HTML]{B7B7B7} 0.28}           & {\color[HTML]{B7B7B7} 72.02}           & {\color[HTML]{B7B7B7} 27.34}           & {\color[HTML]{B7B7B7} -2}          & {\color[HTML]{B7B7B7} 5.83}           &                                       \\ \hline
                        & SDA        & 376.48                                 & 79.82                                  & \cellcolor[HTML]{B6D7A8}\textbf{0.25} & \cellcolor[HTML]{D9EAD3}0.26          & \cellcolor[HTML]{B6D7A8}\textbf{68.21} & 6.83                                   & \cellcolor[HTML]{D9EAD3}2          & 5.50                                  & -                                     \\
                        & MakeItTalk & 256.88                                 & 17.26                                  & 0.02                                  & 0.80                                  & 62.36                                  & 2.07                                   & -3                                 & 3.75                                  & -                                     \\
                        & Wav2Lip*    & \cellcolor[HTML]{D9EAD3}193.32         & \cellcolor[HTML]{D9EAD3}12.57          & 0                                     & -                                     & 46.87                                  & 1.07                                   & \cellcolor[HTML]{D9EAD3}-2         & \cellcolor[HTML]{B6D7A8}\textbf{6.68} & -                                     \\
                        & PC-AVS     & 333.94                                 & 22.53                                  & 0.02                                  & 0.20                                  & \cellcolor[HTML]{D9EAD3}70.36          & \cellcolor[HTML]{D9EAD3}6.93           & -3                                 & \cellcolor[HTML]{D9EAD3}6.17          & -                                     \\
                        & EAMM       & 196.82                                 & 19.40                                  & 0                                     & -                                     & 58.91                                  & 1.65                                   & \cellcolor[HTML]{D9EAD3}-2         & 4.26                                  & -                                     \\
                        & Ours       & \cellcolor[HTML]{B6D7A8}\textbf{88.61} & \cellcolor[HTML]{B6D7A8}\textbf{12.45} & \cellcolor[HTML]{D9EAD3}0.28          & \cellcolor[HTML]{B6D7A8}\textbf{0.36} & 64.30                                  & \cellcolor[HTML]{B6D7A8}\textbf{6.99}  & \cellcolor[HTML]{B6D7A8}\textbf{1} & 4.52                                  & -                                     \\
\multirow{-7}{*}{\rotatebox[origin=c]{90}{CREMA}}   & GT         &                                        &                                        & {\color[HTML]{B7B7B7} 0.24}           & {\color[HTML]{B7B7B7} 0.40}           & {\color[HTML]{B7B7B7} 68.76}           & {\color[HTML]{B7B7B7} 7.76}            & {\color[HTML]{B7B7B7} 1}           & {\color[HTML]{B7B7B7} 5.14}           &
\end{tabular}
\caption{Comparison with other methods. The best scores are in dark green and bold, second bests are in light green. $\uparrow$ / $\downarrow$ indicate higher/lower is better, respectively. Lack of arrow indicates the closer to GT the better. *All of the other methods are one-shot. For fair comparison, Wav2Lip videos were generated using still images, \ie only mouth regions change.}
\label{tab:metrics}
\end{table*}

Unlike other methods \cite{vougioukas2020realistic, prajwal2020lip, chen2020talking, song2022everybody, zhou2021pose, ren2021pirenderer, ji2022eamm}, we do not use any explicit loss function to promote better lip sync of generated samples. Solutions that rely on using dedicated perceptual losses based on pre-trained lipreading models have been effective in improving lip motion accuracy \cite{prajwal2020lip,vougioukas2022generation}. However, Diffused Heads works on frames, not sequences so sequence-based losses can not be applied, and more importantly, during diffusion model training, the goal is to predict the noise that was used on the target frame. Getting back from predicted noise to initial $x_0$, which is required to apply the perceptual loss, is not accurate enough in a single step, and computationally inefficient in more steps.

We introduce a simpler solution: an additional \textit{lip sync loss} $L_{ls}$. During the training, we leverage facial landmarks to crop each frame around the mouth area, and minimize noise prediction in this region:
\begin{equation}
    L_{ls} := \mathbb{E}_{t, x_0, \epsilon} \big[||\Tilde{\epsilon} - \Tilde{\epsilon}_{\theta}(x_t, t) || ^ 2\big]
\end{equation}
where $\Tilde{\epsilon}$ and $\Tilde{\epsilon}_{\theta}$ indicate cropped versions of ground truth and predicted noise, respectively. The process is visualized in Figure \ref{fig:lip_loss}. With the lip loss, the model pays more attention to lip synchronization with audio embeddings, improving the overall perception of sampled videos. We weight $L_{ls}$ with a constant $\lambda_{ls}$ that leverages the model's attention to details of a mouth region and the rest of the frame. We discuss the choice of $\lambda_{ls}$ in Section \ref{sec:ablation}.

The final optimization objective becomes:
\begin{equation}
    L_{simple} + \lambda_{vlb} L_{vlb} + \lambda_{ls} L_{ls}
\end{equation}
where $L_{simple}$ and $L_{vlb}$ are defined by Equations \eqref{eq:l_simple} and \eqref{eq:l_vlb}, respectively.

\subsection{Sampling}
For sampling, only an identity frame and audio embeddings extracted from a speech recording are required. We start video generation by initializing $x_{Motion}^{(0)}$ with copies of the identity frame. Each frame is sampled following the denoising process of diffusion models defined by variational posterior in Equation \eqref{eq:variational_posterior}. After every step, we replace the latest motion frame with a synthesized one. $y_{Motion}^{(k)}$ follows the same procedure as
during the training. 

Generation of a single frame takes a significant amount of time since it requires the model to make a prediction for all of the diffusion time steps $\{1, \dots, T\}$. To speed up the process, methods like DDIM \cite{song2020denoising} or time step respacing can be used. In this work, we use the latter reducing sampling time by a factor of 5.

During experiments, we observed that our model sometimes failed when generating sudden head movements. It synthesizes sequences frame-by-frame, and any occurring errors accumulate in later steps. One of the associated problems is that during training all of the motion frames come from the dataset. Meanwhile, during generation, we use previously sampled frames that have some distortions. We hypothesize that with this setting, the motion frames and the identity frame are equally important in terms of extracting a person's attributes.

To force the model to take more information on the person's appearance from the identity frame, we convert each motion frame to a grayscale. The intuition behind this is that it should make it harder for the model to extract identity features (such as color) while pushing it to seek motion information instead. We found this solution to work well on more complex datasets with a big number of participants.

\section{Experiments}

We evaluate Diffused Heads on the most commonly used datasets for talking face generation: CREMA \cite{cao2014crema} and LRW \cite{chung2016lip}. We compare our method qualitatively and quantitatively to the current state-of-the-art in guided \cite{prajwal2020lip, zhou2020makelttalk, zhou2021pose, ji2022eamm} and pose guidance-free \cite{vougioukas2020realistic} video synthesis. To experience the full quality of our results, readers are strongly encouraged to watch generated videos in the supplementary materials. We will release our code for public use.

\subsection{Implementation Details}
Our model is trained on 128x128 resolution videos. We use the same UNet \cite{ronneberger2015u} architecture as proposed in \cite{dhariwal2021diffusion}, with audio conditioning explained in Section \ref{sec:speech}. We use 256-512-768 channels for the input blocks with 2 ResNet \cite{he2015deep} layers each. In the early stages of experiments, we found adding more attention layers worsened the quality of generated frames. Thus, we only use an attention layer with 4 heads and 64 head channels in the middle block.

\subsection{Qualitative results}

We present qualitative comparison with other methods on CREMA and LRW in Figure \ref{fig:qual_results}. Videos can be found in the supplementary materials. Diffused Heads generates videos that are hard to distinguish from real ones. The faces have natural expressions, eye blinks, and grimaces. The model is able to preserve smooth motion between frames and identity from a given input frame. There are hardly any artifacts, and difficult objects such as hair or glasses are generated accurately. Additionally, Diffused Head works well on challenging videos with people shown from a side view. The important thing to note is that our model does not suffer from mode collapse, and to prove our claim, we will share the entire generated test set on the project's website.

\subsection{Quantitative results}

\begin{table}[]
\centering
\begin{tabular}{cc}
Method                & Score    \\ \hline
PC-AVS \cite{zhou2021pose}    & 34.95\%      \\
Diffused Heads (ours) & 68.72\%  \\
Real videos           & 64.00\%   
\end{tabular}
\caption{Turing test on LRW \cite{chung2016lip} dataset. 10 videos per method and 10 real ones (30 in total) were shown to 140 people. They were asked to vote on whether the samples were real or not. The scores indicate how authentic the videos seemed to the participants.}
\label{tab:turing}
\end{table}

We compare Diffused Heads to other methods: SDA \cite{vougioukas2020realistic}, Wav2Lip \cite{prajwal2020lip}, MakeItTalk \cite{zhou2020makelttalk}, PC-AVS \cite{zhou2021pose}, and EAMM \cite{ji2022eamm}. Results can be found in Table \ref{tab:metrics}. We use first frames and audio sequences from test splits of CREMA and LRW datasets to generate clips. For a fair comparison, the driving videos for PC-AVS and EAMM are chosen randomly from the test sets, and the audio recordings are the same for all of the models.

To measure quality of generated videos, we calculate Fréchet Inception Distance (FID) \cite{heusel2017gans} and Fréchet Video Distance (FVD) \cite{unterthiner2019fvd}. FVD shares the same principles as FID but is extended to video domain, penalizing unnatural dynamics of consecutive frames.

To overcome one of the main challenges of video evaluation, we propose to use the average Optical Flow Magnitude (OFM) of consecutive frames, and Frame-wise Mean Square Error (F-MSE) to measure sequence smoothness (see Appendix for definitions). However, values close to zero are not desirable, since a single repeating image would get a perfect score. Hence, we treat them as population-reference metrics, \ie we want them to be as close as possible to their corresponding ground truth values. Similarly, to evaluate expressiveness, we report population-reference metrics obtained using a blink detector: the average number of blinks per second and the median blink duration.

Lip (WER) and audio-visual (AV Offset, AV Confidence) syncs are measured using a pre-trained lipreader and Syncnet \cite{Chung16a}, respectively. 

Unlike other approaches that mimic head movements from a driving video, our model hallucinates the head motion. As a result, metrics that measure deviation from a ground truth sequence (\eg PSNR, SSIM) heavily penalize our method and are not suitable for evaluation.



Clips generated by our model achieve state-of-the-art scores in FVD and best or close-to-best scores in FID, blinks per second, blink duration, smoothness (OFM and F-MSE), and AV Offset. These metrics show our samples being the best looking and on par with head movements and facial expressions presented in the real videos while maintaining smooth motion and good AV sync.
Worse performance in terms of WER is a consequence of the lack of an expert method supervising lip sync during training. For the same reason, lower values of AV Confidence are observed. However, they do not differ a lot from the ground truth and are still comparable to the best methods. 
It is also worth noting that SDA generates heavily cropped faces without any head movements. 
Additionally, high FID values of Wav2Lip are the effect of animating only mouth region. The remaining part is a fixed ground truth image which is rewarded by FID.
This is reflected in worse FVD scores that capture unrealistic motion - in this case a lack of movement in the majority of a video.
Diffused Heads produces much more realistic and vivid samples. Overall, our model achieves state-of-the-art on most of the metrics.

Moreover, Diffused Heads wins in the most important metric in visual data synthesis - human perception (see Table \ref{tab:turing}). To prove it, we conducted a Turing test on 140 participants. We picked 10 test videos from the LRW dataset generated by the current state-of-the-art method PC-AVS, 10 from our model, and 10 real ones. We sent them to both females and males from different backgrounds and countries. Each of them, after watching every one of the 30 videos, was asked to vote whether they were real or not. 
Diffused Heads performed much better than PC-AVS, and also achieved higher scores than real videos. 
LRW videos contain jitter due to the landmark detection and cropping procedure that was used by the authors of the original paper. This high-frequency noise is not captured by our model leading to smoother sequences (OFM and F-MSE metrics in Table \ref{tab:metrics}) that may seem more natural.


\subsection{Ablation study}
\label{sec:ablation}

\begin{table*}[]
\centering
\begin{tabular}{cccccccc}
\begin{tabular}[c]{@{}c@{}}Motion audio \\ embeddings\end{tabular} & \begin{tabular}[c]{@{}c@{}}Motion\\ grayscale\end{tabular} & CPBD $\uparrow$ & MSE $\downarrow$ & SSIM $\uparrow$ & PSNR $\uparrow$  & LMD $\downarrow$ & WER $\downarrow$ \\ \hline
\multirow{2}{*}{0}                                                 & \ding{55}                                                  & 0.0857          & 1194             & 0.5767          & 18.2912          & 3.0282           & 0.93             \\
                                                                   & \ding{51}                                                  & 0.0858          & 1148             & 0.5890          & 18.5003          & 2.8471           & 0.93             \\ \hline
\multirow{2}{*}{1}                                                 & \ding{55}                                                  & 0.0872          & 1228             & 0.5782          & 18.2309          & 2.8786           & 0.80             \\
                                                                   & \ding{51}                                                  & 0.0856          & 1131             & 0.5996          & 18.6589          & 2.6705           & 0.81             \\ \hline
\multirow{2}{*}{2}                                                 & \ding{55}                                                  & 0.0831          & 1275             & 0.5658          & 17.9912          & 3.0214           & \textbf{0.72}    \\
                                                                   & \ding{51}                                                  & \textbf{0.0925} & \textbf{1025}    & \textbf{0.6225} & \textbf{19.1072} & \textbf{2.5297}  & 0.77             \\ \hline
\multirow{2}{*}{3}                                                 & \ding{55}                                                  & 0.0882          & 1266             & 0.5678          & 17.9945          & 2.9253           & 0.74             \\
                                                                   & \ding{51}                                                  & 0.0882          & 1184             & 0.5851          & 18.4350          & 2.7590           & 0.75            
\end{tabular}
\caption{Ablation study on LRW \cite{chung2016lip} dataset. $\uparrow$ / $\downarrow$ indicate higher/lower is better, respectively.}
\label{tab:ablation_study}
\end{table*}

\begin{figure}[t]
  \centering
    \includegraphics[width=0.90\linewidth]{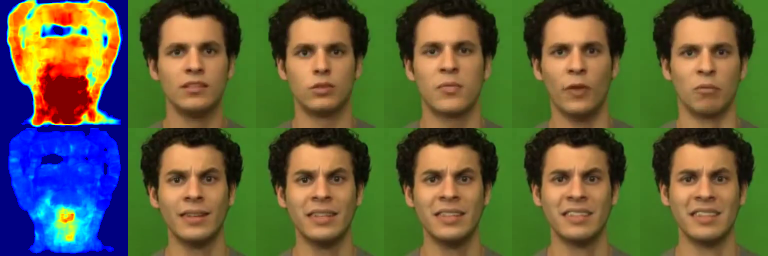}
   \caption{Average magnitudes of optical flow and consecutive frames for 0 (top) and 2 (bottom) motion frames.}
   \label{fig:results_n_frames}
\end{figure}

\begin{figure*}[t]
  \centering
    \includegraphics[width=0.72\linewidth]{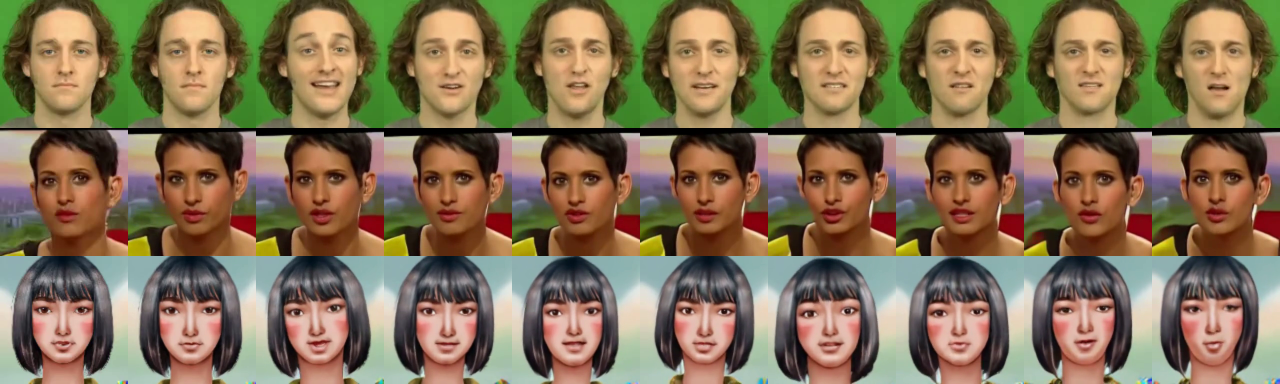}
   \caption{Results of generalization. The audio recordings were (from top): Korean female, German male, and English male. The first row was generated with a model trained on CREMA \cite{cao2014crema}, and the last two with LRW \cite{chung2016lip} one. The first two rows used audio from AVSpeech \cite{ephrat2018looking}, and for the last one, we used a custom image and recording.}
   \label{fig:results_multi}
\end{figure*}

We investigated the influence of a number of motion frames on video quality. We noticed that not using any led to almost random facial expressions. To maintain the motion, we experimented with up to 3 motion frames. 
In Figure \ref{fig:results_n_frames}, we show the comparison between generated videos without any blinks for 0 and 2 motion frames. The average magnitude of an optical flow of a video generated without any motion frames is much higher than the one for 2 motion frames. It is also uniformly high in all of the face regions. It indicates more random movements between consecutive frames. For 2 motion frames, we can spot the highest density around the mouth region, which is the desired behavior.
We include the videos for 0, 1, and 2 motion frames in the supplementary materials where the difference is clearly visible. The model failed to work with 3 motion frames.

For the lip sync loss weight $\lambda_{ls}$, we observed values greater than 0.5 degraded the quality of results. The value of 0.2 gave very realistic results and the best WER score.

Finally, the number of motion audio embeddings and whether to use grayscale on motion frames turned out to be crucial. We present the numerical results of the ablation study on the LRW dataset in Table \ref{tab:ablation_study}. Utilizing grayscale improves the quality of generated videos for every choice of the number of motion audio embeddings. For the latter, the best value to use was 2. 
We noticed that using grayscale on motion frames does not help in less diverse datasets, such as CREMA. It consists of videos of only 91 actors, making generalization to new faces much harder. For that reason, using RGB motion frames lets the model take more information on identity from both identity and motion frames.

\subsection{Generalization}

One of the main challenges in deep learning is the ability of models to generalize well to unseen data. We conducted experiments to show the robustness of Diffused Heads in this manner, and the results can be found in Figure \ref{fig:results_multi}. We show that the model performs well when given part or even all of the input from a different source.

We investigated the behavior of our model with identity frames from CREMA and LRW, and audio recordings from AVSpeech \cite{ephrat2018looking}: a female speaking Korean, and a male German speaker. As the final test, we generated a video of a talking avatar, using an image synthesized by DALL-E 2 \cite{ramesh2022hierarchical} and our own recorded speech. We include more examples in the supplementary video.

The results show Diffused Heads works well on data that comes from outside of a training distribution. The generated frames look pleasant, and the lip movement and facial expressions look natural. Surprisingly, our model was even able to successfully process the image of the avatar, even though it was given only human faces during the training.

\subsection{Limitations}
Despite Diffused Heads achieving state-of-the-art results, it still suffers from some limitations. The main challenge of our method is the length of generated videos. Since we do not provide any additional pose input or visual guidance for head movement and the model autoregressively generates frames, it fails to keep the initial quality for sequences longer than 8-9 seconds. Additionally, diffusion models suffer from long generation times in comparison to other generative models. For now, it is not possible to use our approach in real-time applications, even though it is theoretically suitable for them. New metrics used in talking face generation task are also an open research problem.

\section{Conclusions}
In this work, we presented Diffused Heads: a frame-based method for talking face generation. To synthesize a video that is hard to distinguish by a human from a real one, it only needs one identity frame and an audio sequence containing speech. We evaluated our approach on 2 datasets with different levels of complexity, achieving state-of-the-art results on both of them. We supported this statement by conducting a Turing test on 140 participants showing our results to be indistinguishable from ground truth videos.

{\small
\bibliographystyle{ieee_fullname}
\bibliography{egbib}
}

\newpage
\appendix

\section{Temporal metrics}
We introduce two additional metrics to evaluate the smoothness of generated videos: frame-wise Optical Flow Magnitude (OFM) and Frame-wise Mean Square Error (F-MSE).

Let $X = \{x^{(1)}, \dots, x^{(n)}\}$ be a sequence of frames, and $x^{(k)}_{(i, j)}$ an individual pixel of the $k$-th frame of size $W \times H$. We can define $\text{OFM}(X)$ and $\text{F-MSE}(X)$ as:
\begin{align}
    \text{OFM}(X) &= \frac{1}{WH(n-1)} \sum_{k=2}^n \sum_{i=1}^W \sum_{j=1}^H F\big(x^{(k-1)}_{(i, j)}, x^{(k)}_{(i, j)}\big) \\
    \text{F-MSE}(X) &= \frac{1}{WH(n-1)} \sum_{k=2}^n \sum_{i=1}^W \sum_{j=1}^H \big{\lVert} x^{(k-1)}_{(i, j)} - x^{(k)}_{(i, j)} \big{\rVert}_2^2
\end{align}
where $F$ is an optical flow magnitude between consecutive frames. We use \verb|OpenCV|'s \verb|calcOpticalFlowFarneback| with parameters: \verb|pyr_scale=0.5|, \verb|levels=7|, \verb|winsize=5|, \verb|iterations=15|, \verb|poly_n=5|, and \verb|poly_sigma=1.2|. 

To simplify the notation, we assumed a single channel per pixel. For more channels, the metrics are defined analogically, with additional averaging over channel dimension. Finally, the reported values are averages over all of the videos in the test set. 

The proposed metrics are population-based. We want their values calculated on generated videos to be as close as possible to the ones on ground truth sequences. It is worth noting that a sequence containing just a single repeating frame would get a value of 0 in both of the metrics making them impractical.

\end{document}